%% file: main.tex
\pdfoutput=1
\documentclass[11pt]{article}

\usepackage{EMNLP2023}

\usepackage{times}
\usepackage{latexsym}

\usepackage[T1]{fontenc}
\usepackage[utf8]{inputenc}

\usepackage{microtype}

\usepackage{inconsolata}

\usepackage{times}
\usepackage{latexsym}
\usepackage{subfiles}
\usepackage{graphicx}
\usepackage{booktabs}
\usepackage{array}
\newcolumntype{L}[1]{>{\raggedright\let\newline\\\arraybackslash\hspace{0pt}}m{#1}}
\newcolumntype{C}[1]{>{\centering\let\newline\\\arraybackslash\hspace{0pt}}m{#1}}
\newcolumntype{R}[1]{>{\raggedleft\let\newline\\\arraybackslash\hspace{0pt}}m{#1}}
\title{\textit{Text Encoders Lack Knowledge}: Leveraging Generative LLMs for Domain-Specific Semantic Textual Similarity}  

\author{Joseph Gatto, Omar Sharif, Parker Seegmiller, Philip Bohlman, Sarah Masud Preum \\
Department of Computer Science, Dartmouth College\\
       }
\begin{document}
\maketitle
\begin{abstract}
    Amidst the sharp rise in the evaluation of large language models (LLMs) on various tasks, we find that semantic textual similarity (STS) has been under-explored. In this study, we show that STS can be cast as a text generation problem while maintaining strong performance on multiple STS benchmarks. Additionally, we show generative LLMs significantly outperform existing encoder-based STS models when characterizing the semantic similarity between two texts with complex semantic relationships dependent on \textit{world knowledge}. We validate this claim by evaluating both generative LLMs and existing encoder-based STS models on three newly collected STS challenge sets which require world knowledge in the domains of Health, Politics, and Sports. All newly collected data is sourced from social media content posted after May 2023 to ensure the performance of closed-source models like ChatGPT cannot be credited to memorization. Our results show that, on average, generative LLMs outperform the best encoder-only baselines by an average of 22.3\% on STS tasks requiring world knowledge. Our results suggest generative language models with STS-specific prompting strategies achieve state-of-the-art performance in complex, domain-specific STS tasks. 
\end{abstract}
\begin{figure*}
    \centering
    \includegraphics[width=\textwidth]{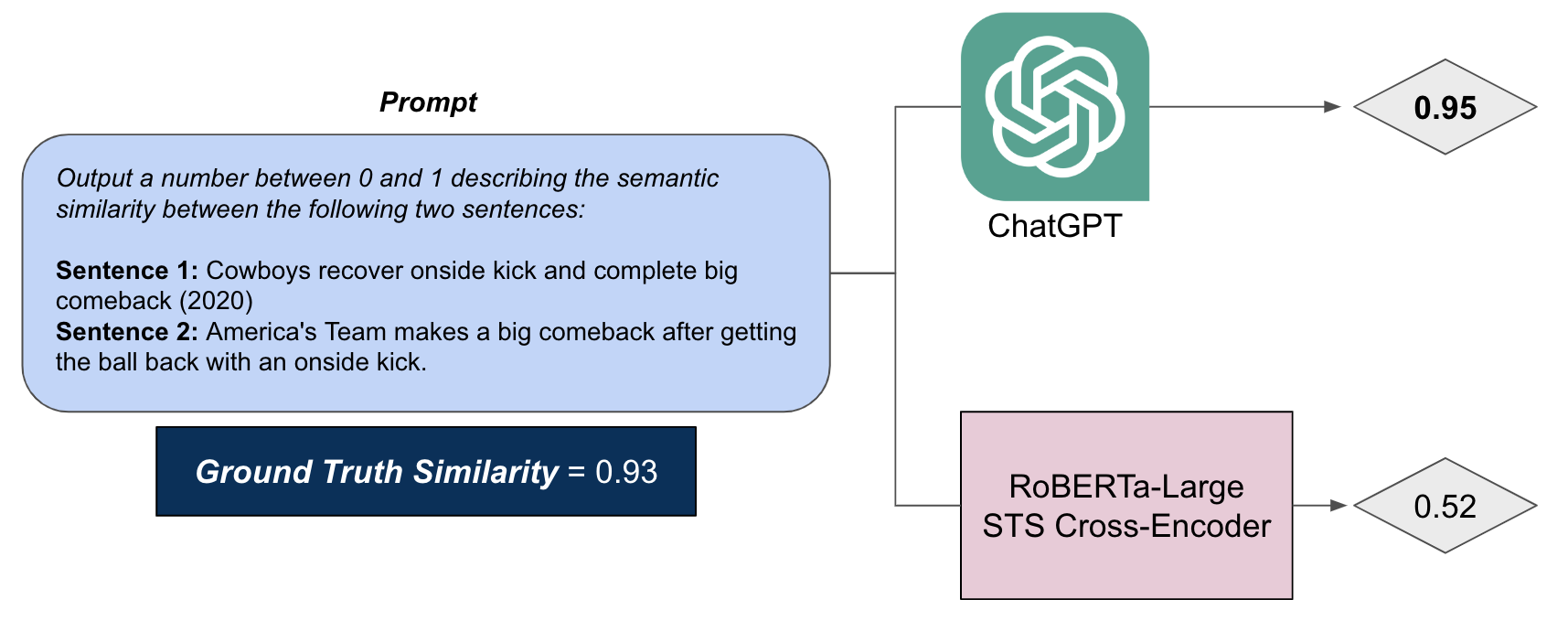}
    \caption{Comparing the performance of ChatGPT vs a RoBERTa-based STS cross encoder on a sample from our STS-Sports challenge set. This sample requires significant world knowledge as proper inference requires knowing 1) that the Cowboys NFL team are often referred to as ``America's Team" and 2) that ``recovering" an onside kick is equivalent to ``getting the ball back" with an onside kick. The prompt corresponds to our best-performing ChatGPT 0-Shot prompt found in Table 2.}
    \label{fig:example_sample} 
\end{figure*}
\subfile{Sections/Introduction}

\subfile{Sections/RelatedWorks}

\subfile{Tables/results_table}
\subfile{Sections/ExperimentalSetup}

\subfile{Sections/Results}

\subfile{Sections/ErrorAnalysis}

\subfile{Sections/Conclusion_Limitations}

\bibliography{anthology,custom}
\bibliographystyle{acl_natbib}
\newpage 
\appendix 
\subfile{Sections/Appendix}

\end{document}

%% file: Sections/Introduction.tex
\section{Introduction}
% \textcolor{blue}{Brief notes from Sarah to restructure the intro}

% The scope of decoder-only LLMs for STS is underexplored. Specifically, STS with challenging semantic comparison &/or domain-specific text when world knowledge is required (?). Give some examples (from para 2) --> Discuss the implications of requiring world knowledge, limited labeled data and how ChatGPT is a good candidate for such tasks. 

% Then talk about existing work (current para 1): but maybe reframe it as less negative?
% Something like limited exploration  in existing research focusing on ChatGPT for STS. For example, paper 1 and 2 did XYZ (positive contribution(s)). However, they overlook ..(point out the specific limitations and connect them to the above-mentioned cases) .

% Then talk about what we are doing in this paper to address the limitation of existing explorations (current para 2,3): 
% \dots \textcolor{blue}{end of notes}

% start what the challenge and papers who  look into the challenge. flip para 1 & 2 

The NLP community has seen a rapid advancement in many areas since the onset of large language models (LLMs) trained using Reinforcement Learning with Human Feedback, including text summarization, machine translation, and problem solving, amongst others \citep{yang2023harnessing}. One area that has not been well explored is the applicability of generative LLMs to Semantic Textual Similarity (STS) tasks.

In recent works, it has been explicitly suggested that LLMs are not well-suited for the STS-B task. In \cite{zhong2023can} they support this claim by showing ChatGPT is inferior to pre-trained RoBERTa models on a small (n=50) set of STS samples. In \cite{yang2023harnessing}, they suggest that STS-B, and more generally regression tasks have ``no use case" in the context of LLMs --- citing the extreme misalignment between LLM training and the prediction of a continuous value. In this study, we aim to show that there are two intuitive reasons as to why \textit{LLMs are highly applicable to Semantic Textual Similarity}. 1) \textbf{World Knowledge: } LLMs do not rely on human-labeled data, allowing them to be exposed to a broad range of world knowledge. Very little human-annotated domain-specific data exists for direct STS training or contrastive learning of sentence embeddings \cite{gao-etal-2021-simcse}, making applications of text encoders to niche domains challenging. Thus, if we can apply LLMs to STS, we may greatly expand the set of problem domains where STS is impactful. 2) \textbf{STS Regression May Align with Language Modeling: } The STS task can be formulated such that the output space is constrained to prediction of a continuous value between [0-1]. Such a formulation reduces the task to outputting similarity as a percentage (e.g. Text A and Text B are 60\% similar). During pre-training, LLMs are very likely to see many texts that use percentages in various contexts, as humans frequently cite percentages in natural language. Thus, when we combine LLMs strong pairwise textual reasoning capabilities with their predisposition to percentages in natural language --- LLMs appear well-suited to the STS task. 

% [Can we find a citation related to LLMs and Regression that helps this point? ]

A limitation of using LLMs for STS is they can be highly expensive and inefficient. For example, STS models are often used in information retrieval, where the goal may be to compare a query text to a large number of documents and then rank the documents based on their similarity to the query \citep{nguyen2016ms}. It may not be viable to leverage generative LLMs for such a task in production, as text generation can suffer from low throughput and high cost. % However, small-scale variants of the information retrieval task are viable for ChatGPT, such as \textit{query-based data collection} (i.e. collecting new samples similar to a prototypical example). 
However, there are many small-scale tasks in academic settings where the poor efficiency of LLMs for STS are often of lesser concern. In the literature, we find small-scale applications of STS in the fields of psychology \citep{marjieh2022predicting}, community question answering \cite{hoogeveen2018detecting}, computational social science \cite{maldeniya2017role}, and propaganda detection \cite{mohtaj-moller-2022-tub} which use generic text encoders for knowledge-intensive/domain-specific problems. In this study, we aim to show that LLMs are more well-suited than generic text encoders for such tasks. 

% Our framework can also make small-scale variants of the information retrieval problem viable for LLMs, such as \textit{query-based data collection} (i.e. collecting new samples similar to a prototypical example).  as LLMs have 1) The capacity to model the relationship between complex multi-sentence texts and 2) The world knowledge required to perform this task in niche domains.  

We confirm our intuition that LLMs like ChatGPT are well-suited to perform STS by conducting the first thorough exploration of STS in the context of text generation. 
% We conclude that \textbf{ChatGPT is a state-of-the-art (SOTA) 0-shot domain-specific STS model.} 
We evaluate two LLMs (i.e., ChatGPT, Llama2) for STS in the context of both existing STS benchmarks and domain-specific STS challenge sets. Our work identifies STS-specific prompting strategies that significantly outperform prompts from prior works \cite{zhong2023can}. Specifically, we find that mapping the original [0-5] similarity scale used in STS benchmarks to be between [0-1] significantly improves performance of LLMs on the STS task. In other words, asking LLMs to infer similarity as a percentage improves performance vs. asking LLMs to utilize an arbitrary scale. See Figure \ref{fig:example_sample} for an example STS prompt used in this study. 

% \textcolor{blue}{Need more details about the prompting strategies: intuition and prompt setup}

% \textcolor{blue}{We should refer to figure 1 / some other STS example in the intro.}

On existing benchmarks, we find that a 0-Shot ChatGPT pipeline provides SOTA performance on the STS13 and STS15 datasets, with near-SOTA performance on STS14 and SICK-R (i.e. 0.45\% and 0.51\% difference in correlation respectively) when compared to unsupervised SOTA models. Given the opaque nature of ChatGPT's training data, we confirm our results are not the result of memorization by collecting 3 new STS challenge datasets using texts written after May 2023 across three domains: health, sports, and politics. We develop each dataset such that similarity is difficult to quantify without significant world knowledge and demonstrate that ChatGPT provides SOTA performance for challenging domain-specific STS. 
% Our results show that ChatGPT with STS-specific prompting strategies outperforms \textit{all} baselines (both supervised and unsupervised STS models) by an average of 23\% on the newly collected data --- verifying our claim that ChatGPT provides SOTA performance for challenging domain-specific STS. 
A summary of our contributions is as follows: 

\begin{itemize}
    
    \item We introduce three new domain-specific STS challenge sets in the domains of Health, Politics, and Sports. We show that ChatGPT outperforms the closest text encoder baseline by an average of 22.3\% on STS challenge sets. 
    
    \item We show that with STS-specific prompting strategies, ChatGPT achieves SOTA performance on two STS benchmark datasets and competitive performance in other datasets when compared to SOTA text encoders. 
    
    \item We analyze errors made by ChatGPT to guide future works on LLMs for STS. 
    % \item We provide a thorough investigation into the errors made by ChatGPT to guide future works on LLMs for STS. 
\end{itemize} 

%% file: Sections/RelatedWorks.tex
\section{Related Work}

\subsection{Supervised STS}
In the supervised setting, STS is commonly evaluated as a part of the GLUE benchmark --- specifically on the STS-B dataset, where texts can be cross-encoded by an LLM and fine-tuned for regression. Supervised STS is largely limited to training on samples sourced from news headlines and image captions --- making such models limited in scope when applied to new domains. LLMs are well-suited to generalize to domain-specific STS data as they contains vast world knowledge. We compare LLMs to both RoBERTa-base and RoBERTa-large \cite{roberta} fine-tuned on the STS-B dataset on our 3 domain-specific datasets. 

%% Taking note of this, will see if other datasets are big enough / applicable 
% \textcolor{blue}{Also is it relevant to compare STS-B to other STS datasets? Although they are not a part of GLUE, it still could be useful to examine how they compare with STS-B.}

\subsection{Unsupervised STS} 
Unsupervised STS occurs when two texts are independently encoded and then compared using measures of embedding similarity. A seminal work in the field of unsupervised STS is SBERT \cite{reimers-gurevych-2019-sentence}, which displays how NLI samples can be used to teach BERT \cite{devlin-etal-2019-bert} how to pool sequences of token embeddings to provide a single vector representation of a given text.  Later improvements on SBERT include SimCSE \cite{gao-etal-2021-simcse} which leveraged contrastive learning to produce better sentence representations. Current state-of-the-art models such as GenSE \cite{chen-etal-2022-generate} produces SOTA results on STS tasks via large-scale synthetic generation of contrastive training triplets.

% Various perspectives may be taken regarding whether or not it is fair to compare ChatGPT to unsupervised STS as unsupervised encoder-only models embed texts separately while ChatGPT encodes the texts together. However, unsupervised STS models have the advantage of being optimized to perform a single task --- while ChatGPT is not designed to perform regression. Thus, while it is not a 1-to-1 comparison, we demonstrate how evaluating ChatGPT in the 0-shot setting on unsupervised STS datasets can inform  ChatGPT's scope for STS. 

LLMs and unsupervised STS use different approaches for text encoding, making their direct comparison difficult. For example, unsupervised STS models excel at this specific task but have fewer parameters, while LLMs are not designed for regression, but have far more parameters and are trained on large-scale unsupervised data. Nonetheless, evaluating LLMs in the 0-shot setting on unsupervised STS datasets can provide insights into their capabilities for STS.

%% file: Tables/results_table.tex
% \begin{table*}[t]
%     \begin{center}
%     \centering
%     \small
% % Change this to be of format xx.xx    
%     \begin{tabular}{lcccccccc}
%     \toprule
%        Model                                     & STS12 & STS13 & STS14 & STS15 & STS16 & STS-B & SICK-R  & Avg. \\
%     \midrule
    
%         SBERT ('all-MiniLM-L6-v2')               & 0.72          & 0.81          & 0.76          & 0.85          & 0.790         & 0.xx & 0.xx & 0.xx \\
%         Sup-SimCSE-BERT-base                     & 0.75          & 0.85          & 0.80          & 0.85          & 0.80          & 0.xx & 0.xx & 0.xx \\
%         Sup-SimCSE-RoBERTa-large                 & 0.77          & 0.87          & 0.82          & 0.87          & 0.840         & 0.xx & 0.xx & 0.xx \\
%         GenSE+                                   & \textbf{0.81} & 0.88          & \textbf{0.85} & 0.89          & \textbf{0.86} & 0.xx & 0.xx & 0.xx \\
%         ChatGPT                                  & 0.69          & \textbf{0.89} & \textbf{0.85}          & \textbf{0.90} & 0.84          & 0.xx & 0.xx & 0.xx \\
%     \end{tabular}
%     % }
%     \end{center}

%     \caption{
%         Results on all popular/standard STS tasks. We find that ChatGPT produces SOTA on 2/5 datasets, 
%     }
%     \vspace{-5pt}
%     \label{table:results}
% \end{table*}

\begin{table*}[!th]
    \centering

\begin{tabular}{lrrrrrrr}
\toprule
                                {} &  STS12          &  STS13          &  STS14          &  STS15         &  STS16            &   STS-B          &  SICK-R \\
\midrule
SBERT                              &  72.37          &  80.60          &  75.59          &  85.39         &  78.99            &  82.03          &   77.15 \\
SimCSE-BERT-B                      &  75.30          &  84.67          &  80.19          &  85.40         &  80.82            &  84.26          &   80.39 \\
SimCSE-RoBERTa-L                   &  77.46          &  87.27          &  82.36          &  86.66         &  83.93            &  86.70          &  \textbf{81.95} \\
GenSE+                             &  \textbf{80.66} &  88.18          &  \textbf{84.69} &  89.03         &  \textbf{85.82}   &  \textbf{87.88} &   80.10 \\
\midrule 
Llama2-7b (Baseline Prompt [0-5])  &  44.05          &  50.27          &  43.03          &  46.02         &  27.23            &      44.37      &        45.33 \\
Llama2-7b (STS Prompt [0-5])       &  42.59          &  41.66          &  30.37          &  33.30         &  26.62            &      35.79      &        39.30 \\
Llama2-7b (STS Prompt [0-1])       &  51.83          &  67.74          &  60.77          &  57.48         &  61.73            &      64.56      &        62.48 \\
\midrule 
ChatGPT (Baseline Prompt [0-5])    &  64.86          &  85.66          &  79.05          &  86.15         &  79.75            &  82.62          & 81.44 \\
ChatGPT (STS Prompt [0-5])         &  64.58          &  86.07          &  80.15          &  85.99         &  79.27            &  81.31          & 78.77 \\
ChatGPT (STS Prompt [0-1])         &  68.97          &  \textbf{89.09}          &  84.24          &  \textbf{89.11}         &  84.54            &  84.73          & 79.84 \\

\bottomrule
\end{tabular}
\caption{Results comparing baseline encoder-only LMs to ChatGPT on standard 7 STS datasets based on Spearman correlation. We find that ChatGPT achieves SOTA results on STS13 and STS15 as well as extremely competitive performance on STS14 and SICK-R. Note: [0-5] prompts use the original similarity score scale of [0.0-5.0]. Our results show that mapping the labels to be between [0.0-1.0] provides a significant performance increase. }
\label{table:results_main}
\end{table*}

%% file: Sections/ExperimentalSetup.tex
\section{Methods}
\subsection{Experimental Setup}

\paragraph{Benchmarking LLMs on 0-Shot STS:} We evaluate various STS-specific 0-shot prompting strategies. An example of our 0-shot inference can be found in Figure \ref{fig:example_sample}. We compare our approach to three baseline unsupervised STS models, which use encoder-only LMs to evaluate sentence representations. Specifically, we explore SBERT\footnote{Huggingface model string: `sentence-transformers/all-MiniLM-L6-v2'}\cite{reimers-gurevych-2019-sentence}, SimCSE \cite{gao-etal-2021-simcse}, and GenSE+ \cite{chen-etal-2022-generate}. 

\paragraph{Domain-Specific STS: } We explore the performance of 0-shot, few-shot, and chain-of-thought (COT) prompting strategies on our domain-specific datasets. Our 0-shot methodology on domain-specific texts follows our best 0-shot prompt as determined by performance on the benchmark STS datasets. For few-shot prompting, we use 5 examples which were manually crafted by the authors. Note, we did no prompt optimization but rather aimed to write a simple prompt that introduced the LLM to the label space as suggested by \cite{min-etal-2022-rethinking}. In each example, we use the same sentence 1, but a different sentence 2, producing evenly spaced similarity scores between 0 and 1, exposing the model to the complete spectrum of label space.  Our COT prompting strategy follows a 1-shot paradigm, showing the model one example of how to reason about the solution step-by-step. The authors wrote the COT example and instructed the model to output the score between a set of brackets (e.g. [semantic similarity = 0.3]) to enable easy prediction extraction. All prompts used in this study can be found in Section \ref{prompting}.

We compare LLMs to both supervised and unsupervised STS models. For supervised models, we use the RoBERTa-base and RoBERTa-large cross-encoders provided by the Sentence-Transformers library\footnote{sbert.net}, which are fine-tuned on the STS-B dataset. 

% \textcolor{blue}{Need more details about CoT prompts and exp. settings for few-shot and CoT here, e.g., sample selection method}

\paragraph{Evaluation Details: } The evaluation pipeline follows \cite{gao-etal-2021-simcse}, which reports the Spearman's rank correlation between all predicted and ground truth similarity scores for all samples in a given dataset. To conduct our experiments, we evaluate two LLMs 1) ChatGPT (`gpt-3.5-turbo-0301') from OpenAI and 2) Llama2-7b \cite{touvron2023llama} from Meta\footnote{Huggingface model string: `Llama-2-7b-chat-hf'}. We choose these two models as they are extremely popular, easy to access, and represent the highest-performing LLMs at their given scales \cite{touvron2023llama}. Note, we exclude GPT-4 from the experimentation due to its significantly higher cost.

We report results after a small grid search on the temperature and top-p hyperparameters of the LLMs. For both models, we use temperature = 0 , top-p = 1. Since Llama2 requires a non-zero temperature, we use 0.0001 as our zero temperature parameter. Additional details regarding our hyperparameter selection can be found in Appendix \ref{hyper}.

\subsection{Extracting Predictions from LLMs}
We use a simple string parsing mechanism to extract predictions from generative LLMs. For 0-Shot and Few-Shot models, we simply return the first number outputted by the model. For COT methods, we extract the decimal found in the set of brackets which the LLM is instructed to produce during inference. If a text cannot be parsed (i.e. no number is output by the model) then we default to a prediction of 0 similarity. 

We note some qualitative analysis regarding the above design choices. First, our highest performing model, ChatGPT, is very good at following STS prompt instructions and thus almost exclusively outputs a single number, so rarely do we default to 0. For lesser-performing models like Llama2, it can happen more frequently, but is still a rare occurrence.

\subsection{Datasets}
\subsubsection{Benchmark Datasets}
Each model is evaluated on the standard 7 STS benchmark datasets: STS 12-16 \cite{agirre-etal-2012-semeval, agirre-etal-2013-sem, agirre-etal-2014-semeval, agirre-etal-2015-semeval, agirre-etal-2016-semeval}, STS-B  \cite{cer-etal-2017-semeval}, and SICK-R \cite{marelli-etal-2014-sick}. All samples in each dataset are annotated on a scale of [0-5], where the mean similarity score across multiple annotators is the final continuous value. 

\subsubsection{Challenge Datasets}
We additionally evaluate each model on 3 newly collected datasets with data collected after May 2023 to ensure ChatGPT's performance is not due to memorization of any information regarding the standard STS benchmarks. Furthermore, this data allows us to evaluate each model's capacity to perform STS when greater world knowledge is required. Our three datasets are \textbf{1) STS-Sports:} Reddit headlines about the National Football League (NFL) and National Basketball Association (NBA); \textbf{2) STS-Health}: Texts sourced from online discussions on Reddit regarding Long COVID; and \textbf{3) STS-News}: A Reddit dataset of recent political headlines. Each dataset has (n=100) text pairs. The data was collected by the authors with the goal of semantic similarity labels being driven by world knowledge relationships. 

Each sample in each dataset consists of 1 real sample from a given source and one human-generated sample. Human-generated texts were written by the authors and crafted to contrast with the source sample in a manner that produces a diverse set of scores across the similarity spectrum. Specifically, high-similarity pairs often employ complex variations of the same information, which require world knowledge, while low-similarity pairs are often constructed to have high token overlap but low semantic similarity, requiring the model to focus deeply on the semantics.  

We chose to manually construct texts as it is extremely difficult to collect samples such as those presented in Figure \ref{fig:example_sample}, where the texts are on the exact same topic but differ drastically in terms of their presentation. Each pair was annotated by three different researchers at the authors' institution and averaged to produce the final similarity score. Each annotator was ensured to be sufficiently knowledgeable about the domain within which they were annotating. The annotation guidelines provided were identical to those released for the STS13 task. The inter-annotator agreement for each dataset can be found in Appendix \ref{dataset-overview} Table \ref{table:IAA}. Please refer to the appendix \ref{dataset-overview} for additional details on data collection, data statistics, and example data.

%% file: Sections/Results.tex
\section{Results}

\subsection{0-Shot STS}
Our 0-shot STS results on benchmark datasets are summarized in Table \ref{table:results_main}. We find that ChatGPT out-performs text encoders on the STS13 and STS15 datasets. Additionally ChatGPT shows competitive performance on STS14, and SICK-R, where there is only a 0.45\% and 0.51\% difference between ChatGPT and the best encoder baseline. We find that the only dataset on which encoder models significantly out-perform ChatGPT is on STS12. This is in part due to the large number of linguistically incoherent texts in STS12. We further discuss the limitations of ChatGPT on certain types of texts in Section \ref{err_analysis}. Llama2, we find, performs poorly on 0-Shot STS on existing benchmarks. This suggests that STS may be an ability emergent at scale for LLMs, as our 7b parameter Llama2 baseline significantly under-performs all other baselines on STS. 
% Another explanation for such a result is that smaller models may need more explicit instruction (e.g. in-context learning) to perform such a task. Evidence of this explanation is found in Table \ref{table:new_data_results}, where Llama2 obtains encoder-level pefroma

We find that the prompts explored in previous works, which prompt ChatGPT to perform STS on the original [0-5] similarity scale, perform significantly worse than when we map the labels between [0-1]. For example, our mapping translates to asking ChatGPT to predict that two texts have 80\% similarity instead of 4/5 similarity. As shown in Table \ref{table:results_main}, ``Baseline Prompt [0-5]" (taken from \cite{zhong2023can}) and ``STS Prompt [0-5]" perform worse on 6/7 tasks, often by a large margin. We find it to be intuitive that LLMs have an easier time understanding and representing semantic similarity as a percentage, as percentages are commonly used to describe various phenomena in a variety of texts (thus making them more likely to appear in LLM training data) unlike comparisons which use a Likert scale.
\subfile{../Tables/results_new_data}

\subsection{Domain-Specific STS}

In Table \ref{table:new_data_results} we see the results of four different model families on our newly collected STS datasets which heavily depend on world knowledge from three different domains. We find that across all domains, ChatGPT performs significantly better than Llama2 as well as both supervised and unsupervised STS models, beating the next closest model by an average of 22.3\%. ChatGPT's competitive performance on the standard STS benchmarks demonstrates it's ability to perform the task, thus it is intuitive that a model with diverse world knowledge should out-perform existing off-the-shelf STS models which contain limited current world knowledge. For example, success on STS-Sports requires a model to know Lebron James plays for the Los Angeles Lakers. STS-News requires the model to know that congresswoman Alexandria Ocasio-Cortez is known as AOC. STS-Health requires the model to know that ``brain fog" is related to ``confusion" and ``lack of focus". This sort of niche knowledge seems  unreasonable for many encoder models to contain --- which is why we argue that ChatGPT is the best option for domain-specific, STS-dependent NLP tasks looking to employ an off-the-shelf model. 

We note that while Llama2 under-performs ChatGPT on all experiments, it does get a significant performance increase in the Few-Shot setting when compared to 0-shot. This may suggest that smaller LLMs require more explicit instruction to perform well on the STS task. Future works may explore STS-specific in-context learning strategies that enable the use of smaller-scale LLMs on this task.

%% file: Tables/results_new_data.tex
% \begin{table*}[t]
%     \begin{center}
%     \centering
%     \small
    
%     \begin{tabular}{lcccc}
%     \toprule
%        Model                                     & STS-NewsA & STS-NewsB & STS-Hoops & Avg. \\
%     \midrule
    
%         SBERT ('all-MiniLM-L6-v2')               & 0.38           & 0.60          & 0.03          & 0.xxx  \\
%         Sup-SimCSE-BERT-base                     & 0.59           & 0.74          & 0.17          & 0.xxx  \\
%         Sup-SimCSE-RoBERTa-large                 & 0.66           & \textbf{0.78} & 0.34          & 0.xxx  \\
%         GenSE+                                   & 0.67           & 0.74          & 0.43          & 0.xxx  \\
%         ChatGPT                                  & \textbf{0.680} & 0.75          & \textbf{0.80} & 0.xxx  \\

%     \end{tabular}
%     % }
%     \end{center}

%     \vspace{-5pt}
%     \label{table:results_new}
% \end{table*}

\begin{table}[!t]
    \centering

\begin{tabular}[width=\columnwidth]{lrrr}
\toprule
Model &  Sports &  News &  Health  \\

 \midrule
        \multicolumn{4}{c}{\it{Unsupervised Models}}\\
 \midrule
SimCSE-R-L                  &        58.87 &        62.47 &          50.98 \\
GenSE+                      &        42.88 &        56.03 &          40.67  \\
\midrule
        \multicolumn{4}{c}{\it{Supervised Models}}\\
 \midrule
 RoBERTa-B                  &        63.17 &        58.29 &          31.56 \\
 RoBERTa-L                  &        63.59 &        65.56 &          50.33\\
 \midrule
        \multicolumn{4}{c}{\it{Llama2 Experiments}}\\
 \midrule
0-Shot              &        47.34 &     44.58 &       37.10  \\
Few-shot            &        66.52 &     58.04 &       46.51   \\
COT                 &        18.73 &     30.98 &       25.55  \\
\midrule
        \multicolumn{4}{c}{\it{ChatGPT Experiments}}\\
 \midrule
0-Shot              &        80.99 &     87.21 &       \textbf{78.11}   \\
Few-shot            &        82.28 &     80.81 &       68.28   \\
COT                 &        \textbf{83.42} &     \textbf{87.74} &       73.71   \\

\bottomrule
\end{tabular}
\caption{
        Results comparing our two best-unsupervised models (i.e., SimCSE-\underline{R}oBERTa-\underline{L}arge and GenSE+) and two RoBERTa models fine-tuned on STS-B to LLMs on our three newly collected domain-specific datasets. We find that ChatGPT outperforms encoder-only models on all tasks by a significant margin. Note: All 0-Shot prompts follow the best 0-shot strategy as determined by results in Table \ref{table:results_main}.
    }
    \label{table:new_data_results}
\end{table}

%% file: Sections/ErrorAnalysis.tex
\section{Where Does ChatGPT Fail on STS?} \label{err_analysis}
In this section, we analyze the top 500 predicted samples from ChatGPT with the largest absolute difference between prediction and ground truth across five  STS datasets in the 0-shot setting (STS 12-16 ). We aim to surface the types of text pairs ill-suited for semantic similarity modeling with ChatGPT. 
% We specifically discuss ChatGPT's struggle with linguistically unacceptable texts and numeric comparisons.  

\subsection{Linguistic Acceptability}

We qualitatively observed that ChatGPT struggles with samples that are syntactically or grammatically incoherent. We validate this claim by running a RoBERTa-base model fine-tuned on the COLA \cite{warstadt2018neural} dataset \footnote{Huggingface model string: `textattack/roberta-base-CoLA'}, which tests if a text is linguistically acceptable. We find that \textbf{34.6\% of highly inaccurate predictions contain a linguistically unacceptable text}. For example, consider the following sample from STS14:\\ \noindent \textbf{Text 1:} what isn 't how what was sold ? \\\textbf{Text 2: }it 's not how it was sold , gb. \\\textbf{Ground Truth Similarity Score:} 0.32\\ 
ChatGPT has very little content or semantics to rely on when analyzing two linguistically unacceptable texts. Thus, it outputs a high similarity score of 0.8 potentially due to token overlap.

To further verify our claim, we evaluate ChatGPT on STS12 in two different contexts --- all samples vs. only text pairs that are both linguistically acceptable. We choose STS12 as it has a high number of linguistically unacceptable texts. We find that on the linguistically acceptable subset (2195/3108 samples in STS12), we get a correlation of 75.95\%, which is a 6.62\% increase in performance compared to evaluation on all samples. 

\subsection{Numeric Reasoning}
It is well-documented that large language models have trouble with numeric reasoning tasks \cite{chen2023need}. In this study, we find that ChatGPT's definition of what constitutes a semantically similar text is not very sensitive to differences in numeric quantities. In other words, ChatGPT commonly gives high semantic equivalence to linguistically similar texts with very different numeric quantities. This is in contrast to the annotation of the STS12-16 benchmarks, where similarity scores can be very sensitive to numeric differences. 

If we assume that samples with numeric quantities in each text require some numeric comparison, we specifically find that, of the top-500 worst predictions made by ChatGPT, \textbf{12.4\% require a numeric comparison}. Consider the following example: \\
\noindent \textbf{Text 1:} 	singapore stocks end up 0.26 percent \\\textbf{Text 2:} singapore stocks end up 0.11 pct \\\textbf{Ground Truth 
 Similarity Score:} 0.4\\ 
ChatGPT is good at recognizing that both texts pertain to Singapore stocks, however ChatGPT's prediction of 0.95 similarity shows little sensitivity to the numeric difference between the texts. Such a prediction by ChatGPT may be considered accurate in different settings, however under the STS12-16 annotation guidelines produced poor results.

%% file: Sections/Conclusion_Limitations.tex
%%% Move to own file %%% 

\section{Conclusion}
In this study, we show that while smaller LLMs like Llama2 struggle on STS, larger models like ChatGPT are highly capable of performing semantic similarity tasks, as it achieves SOTA performance on 2/7 standard STS datasets. We additionally show that ChatGPT is far superior to existing STS models on world knowledge-dependent comparisons --- as ChatGPT outperforms existing models by an average of 22.3\% on domain-specific STS tasks. 
% Finally, we show that ChatGPT struggles to compare linguistically incoherent texts and those which require numeric comparisons. 
In conclusion, ChatGPT shows promising results for domain-specific STS tasks. 
% Such limitations should guide the future use of ChatGPT on STS. 

\section{Limitations}

A limitation of this work is the use of a closed-source model, making it impossible to verify if the model has encountered the data used in our evaluation sets collected prior to September 2021. Also, frequent updates to ChatGPT make it challenging to anticipate how results may change in the future. 

Additionally, our STS solution may not be suitable for large-scale pairwise comparison tasks due to API costs and slow inference speeds. As it stands, our approach is primarily designed for small-scale analysis seeking high-quality outcomes. To demonstrate this, we introduce three new domain-specific challenging STS datasets. The size of the new datasets is limited as it's expensive to scale the annotation process as we want to ensure high-quality data with reliable annotation. However, the number of samples in our domain-specific evaluation sets is on par with other domain-specific STS datasets \cite{souganciouglu2017biosses}.

Finally, we note that we did not do any prompt optimization as a part of this study, which limits the performance potential of our experiments. Future iterations of this work may find that performance can be increased by employing different few shot/COT examples, or by optimizing the problem description. 

\section{Ethical Considerations}
The datasets introduced in this paper collect samples from a total of 6 different subreddits. All of this information was collected manually from the public-facing site. Samples in STS-Sports and STS-News are headlines or texts that are describing public events and thus contain no sensitive information. We note that while samples in STS-Health do contain posts and comments describing personal health experiences, none of the selected samples contain any personally identifying information and are publicly available on the internet. Additionally, this is not human subjects research and thus qualifies for IRB exemption at authors' institution. Reddit was chosen as a data source because it is a suitable platform to collect time-stamped anonymous data in specific domains and on timely topics. However, in the interest of protecting user privacy we plan to provide paraphrased versions of the user-generated samples in STS-Health so that users cannot be identified via internet search of our dataset as suggested in \cite{benton-etal-2017-ethical}.

% A limitation of this work is it's reliance upon a closed-source model. Specifically, we are unable to verify if the model has seen any of the data used in our evaluation sets which were collected before September of 2021 (i.e. the ChatGPT knowledge cutoff). Additionally, we note that ChatGPT is updated semi-frequently and it is difficult to asses how results will change with future updates. 

% % Another limitation pertains to the evaluation set size. 
% It's expensive to scale the annotation process to a large evaluation set as we want to ensure high-quality data with reliable annotation. However, the number of samples in our domain-specific evaluation sets is on par with other domain-specific STS datasets \cite{souganciouglu2017biosses}.
% % our evaluation is still relatively small as this data is expensive to curate. 

% % We hope the impact of the limitation has been minimized by the use of author-collected samples and employment of expert annotators. 

% Finally, as mentioned in the introduction, our STS solution may not be appropriate for large-scale pairwise comparison tasks due to both API costs and slow inference speeds. As currently structured, our approach is limited to small-scale analysis looking for high-quality results.  

%% file: Sections/Appendix.tex
\appendix
\section*{Appendix}

\section{Dataset Overview}
\label{dataset-overview}
\begin{figure}[!h]
    \centering
    \includegraphics[width = \columnwidth]{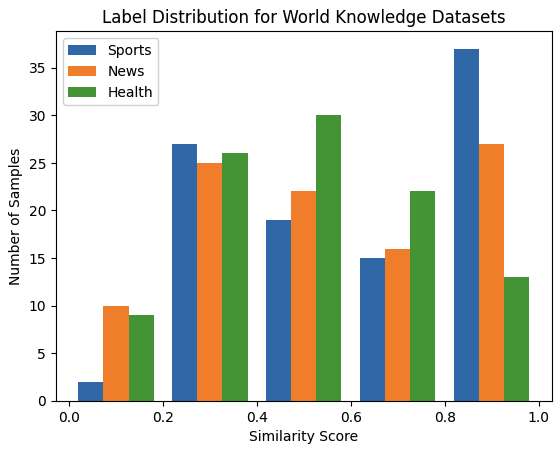}
    \caption{Histogram of the similarity scores for our newly collected sports, news, and health datasets.}
    \label{fig:label_distribution}
\end{figure}

In this section, we provide additional dataset information including sample data and summary statistics of our newly collected datasets. In Figure \ref{fig:label_distribution} we show the distribution of ground truth similarity scores for each of our newly collected datasets.

% In the following sections, we provide additional details and examples from each dataset. 

\begin{table}[!h]
% \small
\centering
\begin{tabular}{@{}ll@{}}
\toprule
\textbf{Dataset} & \textbf{IAA} \\ \midrule
STS-Sports       & 80.19        \\
STS-Health       & 73.38        \\
STS-News         & 82.30        \\ \bottomrule
\end{tabular}%
\caption{Inter-Annotator Agreement (IAA) for each of our newly collected datasets. We define IAA as the mean pearson correlation between all annotators. That is, for our three annotators, we report the mean of $3 \choose 2$ correlations above. }
\label{table:IAA}
\end{table}

\subsection{STS-Sports}

This dataset contains post titles from three different sports subreddits: r/NBA, r/NBATalk, and r/NFL. These subreddits were chosen as they pertain to sports within which our annotators have significant domain knowledge. An example text pair from STS-Sports is shown below:

\begin{quote}
    \textbf{Text 1}: [Highlight] Murray calling his own “BANG” and points at Mike Breen\\
    \textbf{Text 2}: Jamal Murray seen yelling Mike Breen's signature catch phrase after hitting a three \\
     \textbf{Similarity Score}: 0.86 \\
     \textbf{Author Explanation}: This is an extremely difficult STS sample as it requires a model to know who Jamal Murray is (basketball player), who Mike Breen is (basketball announcer), and what Breen's catch-phrase is when people hit a three-point shot ("BANG!"). This is a near semantic match with the difference being that in Text 2 there is no mention of pointing at Mike Breen. \\
     \textbf{Chat GPT Output}: 0.8\\
     \textbf{RoBERTa-large Cross-Encoder}: 0.48
\end{quote}

\subsection{STS-Health}

This dataset consists of post titles, post body content, and comments from two different health-related subreddits: r/covidlonghaulers and r/LongCovid . These subreddits were chosen as they contain health discussions which are user-generated (i.e. non-clinical data) and not overly technical. Validating performance on such data shows ChatGPT's capacity to model social health texts which has many important downstream applications in NLP for public health. An example text pair from STS-Health is shown below. 

\begin{quote}
    \textbf{Text 1}: Drs are Gaslighting me\\
    \textbf{Text 2}: My doctor is making me feel like im crazy!  \\
     \textbf{Similarity Score}: 0.93 \\
     \textbf{Author Explanation}: This sample is difficult as there is little token overlap outside of the mention of doctor. Understanding this sample requires the model to know  modern slang terms such as "Gaslighting". \\
     \textbf{Chat GPT Output}: 0.8 \\
     \textbf{RoBERTa-large Cross-Encoder}: 0.57
\end{quote}

\subsection{STS-News}

This dataset contains post titles from r/Politics. We use this subreddit as the post titles are often headlines containing a diverse array of political figures and phrases that requires significant world knowledge. An example text pair from STS-News is shown below.

\begin{quote}
    \textbf{Text 1}: Montana Republican Lawmaker Suggested She'd Prefer Her Daughter Die By Suicide Than Transition\\
    \textbf{Text 2}: Politician makes insensitive comment towards the transgender community  \\
     \textbf{Similarity Score}: 0.66 \\
     \textbf{Author Explanation}: This is a difficult sample that requires the model to understand a very complex and implicit form of hate speech towards the transgender community. The model is unable to rely on any token overlap between the two texts.\\
     \textbf{Chat GPT Output}: 0.6 \\
     \textbf{RoBERTa-large Cross-Encoder}: 0.41 
\end{quote}

\section{LLM Hyperparameters \& Prompts  }
\subsection{LLM Hyperparameters}
\label{hyper}
For both ChatGPT (gpt-3.5-turbo-0301) and Llama2-7b-chat, we evaluated performance on three different hyperparameter configurations: 

\begin{itemize}
    \item Temperature = 0, Top-P = 1
    \item Temperature = 1, Top-P = 1
    \item Temperature = 0, Top-P = 0.01
\end{itemize}

We identify the best configuration for an experiment (i.e. benchmark STS and domain-specific STS) by averaging the results across all datasets for each set of hyperparameters. Whichever configuration produces the highest average performing experiment (rounded to two decimal places) is chosen as the configuration for all datasets in that experiment.  We find that all experiments found the best performance, on average, from the Temperature = 0, Top-P = 1 configuration. However, it is worth noting that Temperature = 0, Top-P = 0.01 often provided equivalent performance in certain experiments. However, we chose the Top-P = 1 for our configuration as this is the default value provided by the OpenAI API and will thus be a more common configuration for future users.

\subsection{Prompts}
\label{prompting}

In this section, we provide details on the STS prompts used to produce our results. We note that Llama2 struggled to perform the STS task in the 0-shot setting without a specific prompt structure. Specifically, it is the case that the 0-shot prompts in this section all needed to have ``Output:" added to the end of the prompt for the model to properly output its prediction. Thus, in the 0-shot examples that follow, we display the ChatGPT version of the prompt. The Llama2 version is the same prompt with the addition of ``Output:" appended to the end. 

\begin{enumerate}

    \item \textbf{Baseline Prompt [0-5]}\\
    \begin{enumerate}
        \item \textbf{Motivation:} This prompt was used in \cite{zhong2023can} to evaluate ChatGPT on a subset of the STS-B dataset. We run this prompt on all datasets in this study as a baseline  reference. 
        \item \textbf{Prompt:} Determine the similarity between the following two sentences: <Text 1> and <Text 2>. The score should be ranging from 0.0 to 5.0, and can be a decimal.
    \end{enumerate}
    
    \item \textbf{STS Prompt [0-1]}\\
    \begin{enumerate}
        \item \textbf{Motivation:} Our highest performing prompt. We find that having ChatGPT predict labels which are mapped between [0-1] significantly improve performance. 
        \item \textbf{Prompt:} \\
        Output a number between 0 and 1 describing the semantic similiarity between the following two sentences: 

    \underline{Sentence 1}: <Text 1>\\
    \underline{Sentence 2}: <Text 2>\\
          
    \end{enumerate}

    \item \textbf{STS Prompt [0-5]}\\
    \begin{enumerate}
        \item \textbf{Motivation:} To validate our claim that ChatGPT performs better on normalized STS labels, we run the same prompt on the original STS scale of [0-5]. 
        \item \textbf{Prompt:} \\
        Output a number between 0.0 and 5.0 describing the semantic similiarity between the following two sentences: 

    \underline{Sentence 1}: <Text 1>\\
    \underline{Sentence 2}: <Text 2>\\
          
    \end{enumerate}

    \item \textbf{Few Shot STS Prompt [0-1]}: 
    \begin{enumerate}
        \item \textbf{Motivation:} Few shot prompting is a well-established method in the literature \cite{pretrain_prompt}. We thus evaluate on few shot prompting as a baseline measure. Note: The samples used in the few shot prompt were crafted by the authors with the goal of being domain agnostic while introducing the model to the full spectrum of the label space. 

        \item \textbf{Prompt}: 
        \\
        Output a number between 0 and 1 describing the semantic similiarity between the following two sentences: 
    \\
    
        \underline{Sentence 1}: John gave two apples to annie \\
        \underline{Sentence 2}: The ball bounced on the ground \\
        \underline{Similarity Score}: 0 
        \\
        
        \underline{Sentence 1}: John gave two apples to annie \\
        \underline{Sentence 2}: Annie is a girl who likes to read \\
        \underline{Similarity Score}: 0.25
        \\
        
        \underline{Sentence 1}: John gave two apples to annie \\
        \underline{Sentence 2}: Annie likes to eat apples 
        \underline{Similarity Score}: 0.5
        \\
        
        \underline{Sentence 1}: John gave two apples to annie \\
        \underline{Sentence 2}: John gave four apples to annie\\
        \underline{Similarity Score}: 0.75
        \\
        
        \underline{Sentence 1}: John gave two apples to annie \\
        \underline{Sentence 2}: Annie got two apples from john \\
        \underline{Similarity Score}: 1
        \\
        
        \underline{Sentence 1}: <Text 1>\\
        \underline{Sentence 2}: <Text 2> 
    \\
    \end{enumerate}

    \item \textbf{Chain-of-Thought (COT) Prompt:}\\
    \begin{enumerate}
        \item \textbf{Motivation:} Chain-of-Thought prompting has been shown to be a state-of-the-art prompting strategy for many multi-step reasoning tasks \cite{wei2023chainofthought}. We thus evaluate the applicability of COT for STS tasks as a baseline. Note: The 1-shot COT example here was written by the author to avoid interacting with any of the existing STS datasets. 
        \item \textbf{Prompt:} \\
        Discuss how these two texts are similar and different, then assign a semantic similarity score between [0.0-1.0] which describes their semantic similarity:\\ 
        
    \underline{Sentence 1}: Over 50 men have decided that they want to upgrade their iphone\\ 
    \underline{Sentence 2}: We interviewed 25 people and all of them want a new phone \\
    \underline{Similarity}: Lets think step by step. Sentence 1 and Sentence 2 both discuss the upgrade of phones. However they differ in that sentence 1 refers specifically to the iphone and only reports a statistic about men, while sentence 2 discusses phones generally and only for 25 people. Thus, these sentences have a [semantic similarity = 0.7]\\

    Discuss how these two texts are similar and different, then assign a semantic similarity score between [0.0-1.0] which describes their semantic similarity:\\
    
    \underline{Sentence 1}: <Text 1>\\
    \underline{Sentence 2}: <Text 2>\\
    \underline{Similarity}: Lets think step by step.
    \end{enumerate}
    
\end{enumerate}

% \section{Ablation of 0-Shot Prompt Strategy on World Knowledge Datasets}
% % Please add the following required packages to your document preamble:
% % \usepackage{graphicx}
% \begin{table}[!h]
% \setlength{\tabcolsep}{3pt}
% \centering
% \footnotesize
% \begin{tabular}{l|C{1.7cm}C{1.7cm}C{1.8cm}}
% \hline
% \textbf{Dataset} & \textbf{STS Prompt {[}0-1{]}} & \textbf{STS Prompt {[}0-5{]}} & \textbf{Baseline Prompt {[}0-5{]}} \\ \hline
% STS-Sports & 82.12 & 76.45 & 79.62 \\
% STS-Health & 77.67 & 63.92 & 67.45 \\
% STS-News   & 88.22 & 81.77 & 82.51 \\ \hline
% \end{tabular}%
% \caption{Evaluating different 0-shot prompting strategies on our world knowledge datasets.}
% \label{table:ablation}
% \end{table}

% In Table \ref{table:ablation}, we show how all baseline 0-shot prompts perform on our world knowledge datasets. Our results confirm that mapping the labels between [0-1] significantly improves performance over the original [0-5] scale. 